\documentclass[twoside]{article} 
\usepackage{aistats2018}

\usepackage{graphicx}
\usepackage[utf8]{inputenc} 
\usepackage[T1]{fontenc}    
\usepackage{hyperref}       
\usepackage{url}            
\usepackage{booktabs}       
\usepackage{amsfonts}       
\usepackage{nicefrac}       
\usepackage{microtype}      

\usepackage[authoryear]{natbib}
\usepackage{helvet}
\usepackage{courier}
\usepackage{array}
\usepackage{calc}
\usepackage{amsmath}
\usepackage{amssymb}
\usepackage{esint}
\usepackage{paralist}
\usepackage{amsthm}

\usepackage{algorithm}
\usepackage{algorithmic}
\usepackage{caption}
\usepackage{subcaption}

\newtheorem{theorem}{Theorem}

\makeatother

\begin{document}


\twocolumn[

\aistatstitle{Data-Driven Online Decision Making with Costly Observations}

\aistatsauthor{ Onur Atan \And Mihaela van der Schaar }

\aistatsaddress{ Electrical Engineering Department \\ University of California, Los Angeles \\ oatan@ucla.edu \And Department of Engineering Science \\
University of Oxford \\ mihaela.vanderschaar@eng.ox.ac.uk} ]

\begin{abstract}
In most real-world settings such as recommender systems, finance,
and healthcare, collecting useful information is costly and requires
an active choice on the part of the decision maker. The decision-maker
needs to learn simultaneously what observations to make and what actions
to take. This paper incorporates the information acquisition decision into an online learning framework. We propose two different algorithms for this dual learning problem: Sim-OOS and Seq-OOS where observations are made simultaneously and sequentially, respectively. We prove that both algorithms achieve a regret that is sublinear in time.  The developed framework and algorithms can be used in many applications including medical informatics, recommender systems and actionable intelligence in transportation, finance, cyber-security etc., in which collecting information prior to  making decisions is costly. We validate our algorithms in a breast cancer example
setting in which we show substantial performance gains for our proposed
algorithms. 
\end{abstract}
\vspace{-0.2in}
\section{Introduction}
\vspace{-0.1in}

In numerous real-world settings, acquiring useful information is often
costly. In many applications such as recommender systems, finance,
or healthcare, the decision-maker performs costly research/experimentation
to learn valuable information. For instance, a website must pay costs
to observe (e.g. through cookies) the contextual information of its
online users. In doing so it must decide the best information to observe
in order to minimize informational costs while also achieving high
rewards. However, classical contextual Multi-Armed Bandit (MAB) formulations~(\citet{chu2011contextual,slivkins2011contextual,lu2010contextual,dudik2011efficient,langford2007epoch})
have not previously considered these important informational costs
and are thus unable to provide satisfactory performance in such settings. This
paper presents new and powerful methods and algorithms for Contextual
MAB with Costly Observations (CMAB-CO). We show numerically that our
algorithms achieve significant performance gains in breast cancer setting, and we note that the methods and algorithms we develop
are widely applicable (perhaps with some modifications) to an enormous
range of other settings as well, from recommender systems to finance.

A major challenge in these settings is the learning of both optimal
observations and actions. Current MAB methods could potentially be
modified to address this issue by combining the choice of the context
to observe and the action to be taken as a single meta-action and
folding the costs of observations in the rewards. However, the regret
of such an approach can be shown to be exponential in the number of
actions and the number of possible context states; therefore,
it is so inefficient as to be impractical for any realistic problem.
Therefore there is a strong need for the development of new algorithms
that achieve better performance.

To overcome the limitations and challenges discussed above, we propose an alternative approach. We formalize the CMAB-CO problem and show that this problem can be reduced to a two stage Markov Decision Process (MDP) problem with a canonical start state. We propose two different algorithms for this dual learning problem: Sim-OOS and Seq-OOS where observations are made simultaneously and sequentially, respectively. These algorithms build upon the UCRL2 algorithm of~(\citet{jaksch2010near}) to efficiently learn optimal observations and actions.  We show that both Sim-OOS and Seq-OOS algorithms achieve a regret that is sublinear in time. These algorithm thus perform well when the number of observations is small, and it represents a significant improvement over existing algorithms, which would be exponential in the number of observations as well as actions. 

Our main contributions can be summarized as follows: 
\begin{compactenum}
\item We formalize the CMAB-CO problem as a two-stage MDP. 
\item We propose two algorithms under two assumptions: simultaneous and sequential observation selection. We show sublinear in time regret bounds for both algorithms. 
\item We use a breast cancer dataset and show that we can achieve up to significant improvement in performance with respect to an important benchmark. 
\end{compactenum}

As we have noted, our algorithms apply in many settings with different
observations, actions and rewards. In the medical context, the observations
might consist of different types of (costly) medical tests (e.g.,
blood tests, MRI, etc.), actions might consist of choices of treatment,
and rewards might consist of $5$ year survival rates. Hence, an important
aspect of the decision-making is which medical tests to conduct and
which treatment option to recommend. In the recommendation system
context, the observations might consist of (costly) information about
the user (e.g., previous search records, likes in social media, etc.),
actions might consist of item choices and rewards might consist of
click rates. In financial applications, the observations might represent
(costly) research and information gathering about specific assets
(stocks, loans, IPOs, etc.), actions might represent investment decisions,
and rewards might represent investment returns. Indeed, the financial
literature has studied the costs (and incentives) associated with
information gathering in a variety of settings~(\citet{campbel1980information,chemmanur1993pricing}). 
\vspace{-0.1in}
\section{Related Work}
\vspace{-0.1in}
Our paper contributes to multiple strands of literature, including
MAB, MDP and budgeted learning. We describe the contributions of our
work to each topic in turn.
\vspace{-0.1in}
\subsection{MAB Literature}
\vspace{-0.1in}

This work relates to various strands of research in the MAB literature~(\citet{chu2011contextual,slivkins2011contextual,lu2010contextual,dudik2011efficient,langford2007epoch,tekin2014discovering}).
For example, \citet{tekin2014discovering} focuses on learning the
optimal actions by discovering relevant information. However, this
work does not consider the costs associated with gathering information
and is thus unable to provide satisfactory performance in the considered
setting. The CMAB-CO problem is similar to combinatorial semi-bandits
since multiple actions (observations and real actions) are selected
and the rewards of all selected actions (observation cost and real
action rewards) are selected in our setting. However, combinatorial
semi-bandits do not utilize the observed states when taking the action.

Our work is also very related to online probing~(\citet{zolghadr2013online}).
However, the goal in~(\citet{zolghadr2013online}) is to learn the
optimal observations and a single best function that maps observed
features to labels in order to minimize the loss and the observation
cost jointly. Unlike in the considered CMAB-CO setting, an adversarial
setup is assumed and a complete loss feedback (the loss associated
with all the various actions) is obtained at each stage.
\vspace{-0.1in}
\subsection{MDP literature}
\vspace{-0.1in}

The CMAB-CO problem which we consider can be formalized as a two-stage
MDP~(\citet{jaksch2010near,ortner2007logarithmic,osband2014generalization})
with a canonical start state. The action set available in the start
state is the set of observations. Following an observation action
in the start state, the decision-maker moves to a new state (which
consists of the realized states of the selected observations) from
which the decision-maker selects a real action and moves back to the
start state. The reward in the first step is the observation cost
(negative) and the second step is the random reward obtaind by taking
the real action. Stemming from this and building upon the UCRL2 algorithm
of~(\citet{ortner2007logarithmic,jaksch2010near}), we construct
efficient algorithms by exploiting the structure of the CMAB-CO problem:
sparse observation probabilities, known costs.  

\vspace{-0.1in}
\subsection{Budgeted Learning}
\vspace{-0.1in}

The CMAB-CO problem is also similar to budgeted learning as the decision-maker's
goal there is to adaptively choose which features to observe in order
to minimize the loss. For example,~(\citet{cesa2011efficient,hazan2012linear})
adaptively choose the features of the next training example in order
to train a linear regression model while having restricted access
to only a subset of the features. However, these problems do not consider
information costs and are restricted to batch learning.

Another related work is adaptive submodularity~(\citet{golovin2010adaptive})
which aims to maximize rewards by selecting at most $m$ observations/actions.
However, their approach assumes that observation states are statistically
independent and rewards have a submodular structure in observations.

\section{Contextual Multi-armed Bandits with Costly Observations} \label{sec:simulataneous}

\subsection{Problem Formulation}

Next, we present our problem formulation and illustrate it with a
specific example from in the medical context. Let $\mathcal{D}=\{1,2,\ldots,D\}$
be a finite set of observations (types of medical tests such as MRI,
mamogram, ultrasound etc.). Each observation $i\in\mathcal{D}$ is
in a (initially unknown) particular state from a finite set of $\mathcal{X}_{i}$
of possible values (describing the outcomes of the medical tests such
as the BIRADS score associated with a mamogram). Let $\mathcal{X}=\cup_{i\in\mathcal{D}}\mathcal{X}_{i}$
represent the set of all possible state vectors.. The state vector
is $\boldsymbol{\phi}=\left(\phi[1],\phi[2],\ldots,\phi[D]\right),$
where $\phi[i]$ is the state of observation $i$, which represents
the context in the CMAB formulation. We assume that the state vector
is drawn according to a fixed but unknown distribution. We write $\boldsymbol{\Phi}$
to denote a random state vector and $p(\boldsymbol{\phi})=\Pr(\boldsymbol{\Phi}=\boldsymbol{\phi})$
to denote the probability of state vector $\boldsymbol{\phi}$ being
drawn. In the medical context, $p(\cdot)$ models a joint probability
over the results of the medical tests.

We assume that only the states of the observations that are selected
by the decision-maker are revealed in each time instance. Let $\boldsymbol{\boldsymbol{\psi}}$
denote a partial state vector, which only contains the state of a
subset of the selected observations. For example, for selected observations
$\mathcal{I}\subseteq\mathcal{D}$, the partial state vector is $\boldsymbol{\boldsymbol{\psi}}=\left(\boldsymbol{\psi}[1],\boldsymbol{\psi}[2],\ldots,\boldsymbol{\psi}[D]\right)$
with 
\[
\boldsymbol{\psi}[i]=\begin{cases}
\phi[i] & \quad\text{if }i\in\mathcal{I}\\
? & \quad\text{if }i\notin\mathcal{I}
\end{cases}
\]
where $?$ denotes our symbol for missing observation states. We use
the notation $\operatorname{dom}(\boldsymbol{\psi})=\{i\in\mathcal{D}:\boldsymbol{\psi}[i]\neq?\}$
to refer to the domain of $\boldsymbol{\psi}$ (i.e., the set of the
medical test outcomes realized in $\boldsymbol{\psi}$). Let $\Psi^{+}(\mathcal{I})=\{\boldsymbol{\psi}:\operatorname{dom}(\boldsymbol{\psi})=\mathcal{I}\}$
denote the set of all possible partial state vectors with observations
from $\mathcal{I}$ (i.e., the set of all possible medical test outcomes
of $\mathcal{I}$). Let $\Psi=\cup_{\mathcal{I}\subseteq\mathcal{D}}\Psi^{+}(\mathcal{I})$
denote the set of all possible partial state vector states.
We say $\boldsymbol{\boldsymbol{\psi}}$ is \textit{consistent} with
$\boldsymbol{\phi}$ if they are equal everywhere in the domain of
$\boldsymbol{\boldsymbol{\psi}}$, i.e., $\boldsymbol{\psi}[i]=\phi[i]$
for all $i\in\operatorname{dom}(\boldsymbol{\boldsymbol{\psi}})$.
In this case, we write $\boldsymbol{\phi}\sim\boldsymbol{\boldsymbol{\psi}}$.
If $\boldsymbol{\boldsymbol{\psi}}$ and $\boldsymbol{\boldsymbol{\psi}}'$
are both consistent with some $\boldsymbol{\phi}$, and $\operatorname{dom}(\boldsymbol{\boldsymbol{\psi}})\subseteq\operatorname{dom}(\boldsymbol{\boldsymbol{\psi}}')$,
we say $\boldsymbol{\boldsymbol{\psi}}$ is a \textit{substate}
of $\boldsymbol{\boldsymbol{\psi}}'$. In this case, we write $\boldsymbol{\boldsymbol{\psi}}'\succeq\boldsymbol{\boldsymbol{\psi}}$.

We illustrate these definitions on a simple example. Let $\boldsymbol{\phi}=(-1,1,1)$
be a state vector, and $\boldsymbol{\boldsymbol{\psi}}_{1}=(-1,?,-1)$
and $\boldsymbol{\boldsymbol{\psi}}_{2}=(-1,?,?)$ be partial state
vectors. Then, all of the following claims are true: 
\[
\boldsymbol{\phi}\sim\boldsymbol{\boldsymbol{\psi}}_{2},\;\boldsymbol{\boldsymbol{\psi}}_{1}\succeq\boldsymbol{\boldsymbol{\psi}}_{2},\;\operatorname{dom}(\boldsymbol{\boldsymbol{\psi}}_{1})=\{1,3\}.
\]
We consider a MAB setting with costly observations where the following
sequence of the events is taking place at each time $t$: 
\begin{compactenum}
\item The environment draws a state vector $\boldsymbol{\phi}_{t}$ according
to unknown distribution $p(\cdot)$. The state vector is initially
unknown to the decision-maker. 
\item The decision-maker is allowed to select at most $m$ observation at
time $t$, denoted as $\mathcal{I}_{t}$, with paying a known cost
of $c_{i}\in[0,1]$ for each observations $i$ in the set $\mathcal{I}_{t}$.
We assume that the decision-maker has an upper bound $m$ on the maximum
number of observations that can be made at each time $t$. Let $\mathcal{P}_{\leq m}(\mathcal{D})$
denote the subset of the observations with cardinality less than $m$,
i.e., $\mathcal{P}_{\leq m}(\mathcal{D})=\{\mathcal{I}\subseteq\mathcal{D}:|\mathcal{I}|\leq m\}$.
The partial state vector $\boldsymbol{\boldsymbol{\psi}}_{t}$ from
the observations $\mathcal{I}_{t}$ is revealed to the decision-maker,
while the remainder of the states remain unknown to the decision-maker. 
\item Based on its available information $\boldsymbol{\psi}_{t}$, the decision-maker
takes an action $a_{t}$ from a finite set of actions $\mathcal{A}=\{1,2,\ldots,A\}$
and observes a random reward $r_{t}$ with support $[0,1]$ and $\mathbb{E}\left[r_{t}\right]=\bar{r}(a_{t},\boldsymbol{\phi}_{t})$
where $\bar{r}:\mathcal{A}\times\mathcal{X}\rightarrow\left[0,1\right]$
is an unknown expected reward function. 
\end{compactenum}

We overload the definition of $p$ and $\bar{r}$ to denote marginal
probabilities and expected rewards of partial state vectors. We write
$p(\boldsymbol{\boldsymbol{\psi}})=\Pr(\boldsymbol{\Phi}\sim\boldsymbol{\boldsymbol{\psi}})$
to denote the marginal probability of $\boldsymbol{\boldsymbol{\psi}}$
being realized and $\bar{r}(a,\boldsymbol{\boldsymbol{\psi}})=\mathbb{E}\left[\bar{r}(a,\boldsymbol{\Phi})|\boldsymbol{\Phi}\sim\boldsymbol{\boldsymbol{\psi}}\right]$
to denote the marginal expected reward of action $a$ when the partial
state vector is $\boldsymbol{\boldsymbol{\psi}}$. Observe that $\sum_{\boldsymbol{\psi}\in\Psi^{+}(\mathcal{I})}p(\boldsymbol{\boldsymbol{\psi}})=1$.

The \textit{policy} $\pi$ for selecting observations and associated
actions consists of a set of observations $\mathcal{I}$ and an adaptive
action strategy $h:\Psi^{+}(\mathcal{I})\rightarrow\mathcal{A}$,
which maps each possible partial state vectors from $\mathcal{I}$
to actions (e.g., a policy consists of a subset of medical tests $\mathcal{I}$
and treatment recommendation for each possible test results from $\mathcal{I}$).
The expected gain of the policy $\pi=\{\mathcal{I},h\}$ is given
by 
\begin{equation}
\rho(\pi)=\beta\sum_{\boldsymbol{\boldsymbol{\psi}}\in\Psi^{+}(\mathcal{I})}p(\boldsymbol{\boldsymbol{\psi}})\bar{r}(h(\boldsymbol{\boldsymbol{\psi}}),\boldsymbol{\boldsymbol{\psi}})-\sum_{i\in\mathcal{I}}c_{i},
\end{equation}
where $\beta>1$ is the gain parameter, which balances the trade-off
between the rewards and observation costs. For example, $\beta$ represents
the revenue made by one click in the recommendation system context.
The expected gain of the policy $\pi$ is the expected reward of $\pi$
minus the observation cost incurred by $\pi$. Without loss of generality,
we assume that decision-maker is allowed to make at most $m$ observations.
Let $\Pi$ denote the set of all possible policies. The oracle policy
is given by $\pi_{m}^{*}=\operatorname{\arg\max}_{\pi=(\mathcal{I},h)\in\Pi:|\mathcal{I}|\leq m}\rho(\pi)$.

The expected gain of the oracle policy is given by $\rho_{m}^{*}=\rho(\pi_{m}^{*})$.
Note that our oracle is different than the oracle used in the contextual
bandit literature. To illustrate the difference, define $\bar{r}^{*}(\boldsymbol{\psi})=\bar{r}(a^{*}(\boldsymbol{\psi}),\boldsymbol{\psi})=\max_{a\in\mathcal{A}}\bar{r}(a,\boldsymbol{\psi})$
to be the expected reward of the best action when the partial state
vector is $\boldsymbol{\psi}$. We refer to the policy that selects
observations $\mathcal{I}$ and the best actions $a^{*}(\boldsymbol{\psi})$
for all $\boldsymbol{\psi}\in\Psi^{+}(\mathcal{I})$ as the fixed
$\mathcal{I}$-oracle policy. The expected reward of the fixed $\mathcal{I}$-oracle
policy is given by 
\[
V(\mathcal{I})=\beta\sum_{\boldsymbol{\psi}\in\Psi^{+}(\mathcal{I})}p(\boldsymbol{\psi})\bar{r}^{*}(\boldsymbol{\psi})-\sum_{i\in\mathcal{I}}c_{i}.
\]
It can be shown that the oracle policy $\pi_{m}^{*}=(\mathcal{I}_{m}^{*},h^{*})$
is given by 
$h^{*}(\boldsymbol{\psi})=\operatorname{\arg\max}_{a\in\mathcal{A}}\bar{r}(a,\boldsymbol{\psi})$
and 
$
\mathcal{I}_{m}^{*}=\operatorname{\arg\max}_{\mathcal{I}\in\mathcal{P}_{\leq m}(\mathcal{D})}V(\mathcal{I}).
$. Note that $\rho_{m}^{*}=V(\mathcal{I}_{m}^{*})$. Therefore, the oracle
defined in our setting achieves the best expected reward among all
the fixed $\mathcal{I}$-oracle policies.

Consider an adaptive policy $\pi_{1:T}=\left[\mathcal{I}_{t},h_{t}\right]_{t=1}^{T}$,
which takes observation-action $\mathcal{I}_{t}$, observes $\boldsymbol{\psi}_{t}$,
uses this observation to take an action $a_{t}=h_{t}(\boldsymbol{\psi}_{t})$
and receives the reward of $r_{t}$. The cumulative reward of $\pi_{1:T}$
is $\sum_{t=1}^{T}\left(\beta r_{t}-\sum_{i\in\mathcal{I}_{t}}c_{i}\right)$.
The $T$-time regret of the policy $\pi_{1:T}=\left[\mathcal{I}_{t},h_{t}\right]_{t=1}^{T}$
is given by 
\begin{equation}
\operatorname{Reg}_{T}^{\pi_{1:T}}=T\rho_{m}^{*}-\sum_{t=1}^{T}\left(\beta r_{t}-\sum_{i\in\mathcal{I}_{t}}c_{i}\right).\notag
\end{equation}
The goal here is to compute the policy $\pi_{1:T}$ to minimize this
regret by selecting at most $m$ observations. 

Current online learning methods could be modified to address the CMAB-CO
problem by defining a set of \textit{meta-actions} that comprises
all the combinations of observation subsets and actions taken based
on these observations, and then applying a standard MAB algorithm
(such as the UCB algorithm~\citet{auer2002a}) by considering these
meta-actions to be the action space. While this algorithm is straightforward
to implement, it scales linearly with the total number of policies
$|\Pi|=\sum_{\mathcal{I}\in\mathcal{P}_{\leq m}(\mathcal{D})}A^{|\Psi^{+}(\mathcal{I})|}$. This is exponential in the number of state vectors.  This makes such algorithms computationally infeasible and suboptimal (compared to the lower bound) even when the numbers of actions and partial states is small. This poor scaling performance is due to the fact that the algorithm does not take into account that selecting an action yields information for many policies.

\begin{algorithm}[t]
\protect\caption{Simultaneous Optimistic Observation Selection (Sim-OOS) }
\vspace{-0.05in}
\label{alg:PSim-OOS} \begin{algorithmic} \STATE \textbf{Input:}
$m,\left[c_{i}\right]_{i\in\mathcal{D}}$, $\operatorname{conf}_{1}(n,t)$,
$\operatorname{conf}_{2}(n,t),\beta$ \STATE \textbf{Initialize:}
$\mathcal{E}(\operatorname{dom}(\boldsymbol{\psi}),\boldsymbol{\psi})\leftarrow\emptyset$
for all $\boldsymbol{\psi}\in\Psi$. \STATE \textbf{Initialize:}
$\mathcal{E}(\mathcal{I})\leftarrow\emptyset$ for all $\mathcal{I}\in\mathcal{P}_{\leq m}(\mathcal{D})$.
\STATE \textbf{Initialize:} $\mathcal{E}(a,\boldsymbol{\psi})\leftarrow\emptyset$
for all $a\in\mathcal{A}$ and $\boldsymbol{\psi}\in\Psi$. \FOR{rounds
$k=1,2,\ldots$} \STATE $\operatorname{conf}_{1,k}(a,\boldsymbol{\psi})\leftarrow\operatorname{conf}_{1}(N_{k}(a,\boldsymbol{\psi}),t_{k})$.
\STATE $\operatorname{conf}_{2,k}(\mathcal{I})\leftarrow\operatorname{conf}_{2,k}(N_{k}(\mathcal{I}),t_{k})$.
\STATE $\widehat{r}_{k}(a,\boldsymbol{\psi})=\frac{1}{N_{k}(a,\boldsymbol{\psi})}\sum_{\tau\in\mathcal{E}_{k}(a,\boldsymbol{\psi})}r_{\tau}$
for all $a\in\mathcal{A}$ and $\boldsymbol{\psi}\in\Psi$. \STATE
$\widehat{p}_{k}(\boldsymbol{\psi})=\frac{N_{k}(\operatorname{dom}(\boldsymbol{\psi}),\boldsymbol{\psi})}{N_{k}(\operatorname{dom}(\boldsymbol{\psi}))}$
for all $\boldsymbol{\psi}\in\Psi$. \STATE $\widehat{h}_{k}(\boldsymbol{\psi})\leftarrow\operatorname{\arg\max}_{a\in\mathcal{A}}\;\widehat{r}_{k}(a,\boldsymbol{\psi})+\operatorname{conf}_{1,k}(a,\boldsymbol{\psi})$
\STATE Solve the convex optimization problem given in (\ref{eqn:optimization2})
for all $\mathcal{I}\in\mathcal{P}_{\leq m}(\mathcal{D})$ \STATE
Set $\widehat{V}_{k}(\mathcal{I})$ as the maximizer. \STATE $\widehat{\mathcal{I}}_{k}\leftarrow\arg\max_{\mathcal{I}\in\mathcal{P}_{\leq m}(\mathcal{D})}\widehat{V}_{k}(\mathcal{I})$.
\STATE $\nu_{k}(a,\boldsymbol{\psi})\leftarrow0$ for all $a$ and
$\boldsymbol{\psi}\in\Psi$. \WHILE{$\forall(a,\boldsymbol{\psi}):\nu_{k}(a,\boldsymbol{\psi})<\max(1,N_{k}(a,\boldsymbol{\psi}))$}
\STATE Select observations $\widehat{\mathcal{I}}_{k}$, observe
the partial state vector $\boldsymbol{\psi}_{t}$, \STATE Select
action $a_{t}=\widehat{h}_{k}(\boldsymbol{\psi}_{t})$, observe reward
$r_{t}$. \STATE Update $\nu_{k}(a_{t},\boldsymbol{\psi}_{t})\leftarrow\nu_{k}(a_{t},\boldsymbol{\psi}_{t})+1$.
\FOR{$\boldsymbol{\psi}:\boldsymbol{\psi}_{t}\succeq\boldsymbol{\psi}$}
\STATE $\mathcal{E}(\operatorname{dom}(\boldsymbol{\psi}),\boldsymbol{\psi})\leftarrow\mathcal{E}_{k+1}(\boldsymbol{\psi},\operatorname{dom}(\boldsymbol{\psi}))\cup t$.
\STATE $\mathcal{E}(\operatorname{dom}(\boldsymbol{\psi}))\leftarrow\mathcal{E}_{k+1}(\operatorname{dom}(\boldsymbol{\psi}))\cup t$.
\ENDFOR \STATE $\mathcal{E}(a_{t},\boldsymbol{\psi}_{t})\leftarrow\mathcal{E}_{k+1}(a,\boldsymbol{\psi})\cup t$.
\STATE $t\leftarrow t+1$. \ENDWHILE \ENDFOR \end{algorithmic} 
\end{algorithm}
\vspace{-0.1in}

\subsection{Simultaneous Optimistic Observation Selection (Sim-OOS) Algorithm}

To address the above mentioned limitations of such MAB algorithms,
we develop a new algorithm, which we refer to as Simultaneous Optimistic Observation
Selection (Sim-OOS). Sim-OOS operates in rounds $k=1,2,\ldots$. Let $t_{k}$
denote time at the beginning of round $k$. The decision-maker keeps
track of the estimates of the mean rewards and the observation probabilities.
Note that when the partial state vector $\boldsymbol{\psi}_{t}$ from
observation set $\mathcal{I}_{t}$ is revealed, the decision-maker
can use this information to not only update the observation probability
estimate of $\boldsymbol{\psi}_{t}$ but also update the observation
probability estimate of all substates of $\boldsymbol{\psi}_{t}$.
However, the decision-maker cannot update the mean reward estimate
of pairs of $a_{t}$ and substates of $\boldsymbol{\psi}_{t}$
since this would result in a bias on the mean reward estimates. Therefore,
at each round $k$, we define $\mathcal{E}_{k}(a,\boldsymbol{\psi})=\{\tau<t_{k}:a_{\tau}=a,\boldsymbol{\psi}_{\tau}=\boldsymbol{\psi}\}$,
$\mathcal{E}_{k}(\mathcal{I})=\{\tau<t_{k}:\mathcal{I}\subseteq\mathcal{I}_{\tau}\}$
and $\mathcal{E}_{k}(\boldsymbol{\psi},\mathcal{I})=\{\tau<t_{k}:\mathcal{I}\subseteq\mathcal{I}_{\tau},\boldsymbol{\psi}_{\tau}\succeq\boldsymbol{\psi}\}$
if $\boldsymbol{\psi}\in\Psi^{+}(\mathcal{I})$ and $\mathcal{E}_{k}(\boldsymbol{\psi},\mathcal{I})=\emptyset$
if $\boldsymbol{\psi}\notin\Psi^{+}(\mathcal{I})$.

We define the following counters: $N_{k}(\mathcal{I},\boldsymbol{\psi})=|\mathcal{E}_{k}(\mathcal{I},\boldsymbol{\psi})|$,
$N_{k}(\mathcal{I})=|\mathcal{E}_{k}(\mathcal{I})|$, $N_{k}(a,\boldsymbol{\psi})=|\mathcal{E}_{k}(a,\boldsymbol{\psi})|$.
In addition to these counters, we also keep counters of partial state-action
pair visits in a specific round $k$. Let $\nu_{k}(a,\boldsymbol{\psi})$
denote the number of times action $a$ is taken when partial state
$\boldsymbol{\psi}$ is observed in round $k$. Furthermore, we can
express the mean reward estimate and observation probability estimates
as follows: 
\[
\widehat{r}_{k}(a,\boldsymbol{\psi})=\frac{1}{N_{k}(a,\boldsymbol{\psi})}\sum_{\tau\in\mathcal{E}_{k}(a,\boldsymbol{\psi})}r_{\tau},
\]
\[
\widehat{p}_{k}(\boldsymbol{\psi})=\frac{N_{k}(\operatorname{dom}(\boldsymbol{\psi}),\boldsymbol{\psi})}{N_{k}(\operatorname{dom}(\boldsymbol{\psi}))}
\]
provided that $N_{k}(a,\boldsymbol{\psi})>0$ and $N_{k}(\operatorname{dom}(\boldsymbol{\psi}))>0$.
Since these estimates can deviate from their true mean values, we
need to add appropriate confidence intervals when optimizing the policy.
In the beginning of each round $k$, the Sim-OOS computes the policy of
round $k$ by solving an optimization problem given in (\ref{eqn:optimization1}).
The optimization problem with the mean reward estimate and observation
probability estimates is given by 
\begin{align}
 & \underset{\pi=\{\mathcal{I},h\},\tilde{p},\tilde{r}}{\operatorname{maximize}}\;\;\beta\sum_{\boldsymbol{\psi}\in\Psi^{+}(\mathcal{I})}\tilde{p}(\boldsymbol{\psi})\tilde{r}(h(\boldsymbol{\psi}),\boldsymbol{\psi})-\sum_{i\in\mathcal{I}}c_{i}\notag\\
 & \text{{\it subject to}}\;|\tilde{r}(a,\boldsymbol{\psi})-\widehat{r}_{k}(a,\boldsymbol{\psi})|\leq\text{conf}_{1,k}(a,\boldsymbol{\psi}),\;\;\forall(a,\boldsymbol{\psi}),\notag\\
 & \sum_{\boldsymbol{\psi}\in\Psi^{+}(\mathcal{I})}|\tilde{p}(\boldsymbol{\psi})-\widehat{p}_{k}(\boldsymbol{\psi})|\leq\text{conf}_{2,k}(\mathcal{I}),\notag\\
 & \sum_{\boldsymbol{\psi}\in\Psi^{+}(\mathcal{I})}\tilde{p}(\boldsymbol{\psi})=1,\;\;\forall\mathcal{I}\in\mathcal{P}_{\leq m}(\mathcal{D}),\label{eqn:optimization1}
\end{align}
where $\text{conf}_{1,k}(a,\boldsymbol{\psi})$ and $\text{conf}_{2,k}(\mathcal{I})$
are the confidence bounds on the estimators at time $t_{k}$. We will
set these confidence bounds later in order to achieve provable regret
guarantees with high probability. Let $\widehat{\pi}_{k}=\{\widehat{\mathcal{I}}_{k},\widehat{h}_{k}\}$
denote the policy computed by the Sim-OOS.

The Sim-OOS follows policy $\widehat{\pi}_{k}$ in round $k$. At time
$t$ in round $k$ ($t_{k}\leq t\leq t_{k+1}$), the Sim-OOS selects $\widehat{\mathcal{I}}_{k}$
and observes the partial state vector $\boldsymbol{\psi}_{t}$ from
observations $\mathcal{I}_{k}$ and on the basis of this, it takes
an action $\widehat{h}_{k}(\boldsymbol{\psi}_{t})$. Round $k$ ends
when one of the visits to the partial state vector-action pair in
round $k$ is the same as $N_{k}(a,\boldsymbol{\psi})$ (the total
observations of the partial state-action pair from previous
rounds $k'=1,\ldots,k-1$). This ensures that the optimization problem
given in (\ref{eqn:optimization1}) is only solved when the estimates
and confidence bounds are improved. 

The optimization problem in (\ref{eqn:optimization1}) can be reduced
to a set of convex optimization problems which can be solved efficiently
in polynomial time complexity~(\cite{boyd2004convex}) (the details
of this reduction are discussed in the supplementary material). In
round $k$, let $\widehat{r}_{k}^{*}(\boldsymbol{\psi})=\max_{a\in\mathcal{A}}\;\widehat{r}_{k}(a,\boldsymbol{\psi})+\operatorname{conf}_{1,k}(a,\boldsymbol{\psi})$
be the optimistic reward of value of the partial state vector $\boldsymbol{\psi}$
in round of $k$. The optimistic gain of a fixed $\mathcal{I}$-oracle
in round $k$, denoted by $\widehat{V}_{k}(\mathcal{I})$, is defined
as the maximizer of the following optimization problem: 
\begin{align}
 & \underset{\left[\tilde{p}(\boldsymbol{\psi})\right]_{\boldsymbol{\psi}\in\Psi^{+}(\mathcal{I})}}{\text{maximize}}\;\;\;\beta\sum_{\boldsymbol{\psi}\in\Psi^{+}(\mathcal{I})}\tilde{p}(\boldsymbol{\psi})\widehat{r}_{k}^{*}(\boldsymbol{\psi})-\sum_{i\in\mathcal{I}}c_{i}\notag\\
 & \textit{subject to}\;\;\sum_{\boldsymbol{\psi}\in\Psi^{+}(\mathcal{I})}\left|\tilde{p}(\boldsymbol{\psi})-\widehat{p}_{k}(\boldsymbol{\psi})\right|\leq\text{conf}_{2,k}(\mathcal{I}),\notag\\
 & \sum_{\boldsymbol{\psi}\in\Psi^{+}(\mathcal{I})}\tilde{p}(\boldsymbol{\psi})=1.\label{eqn:optimization2}
\end{align}
At any time $t$ of round $k$, it can be shown that the optimization
in (\ref{eqn:optimization1}) can be solved as: $\widehat{h}_{k}(\boldsymbol{\psi})=\operatorname{\arg\max}_{a\in\mathcal{A}}\;\widehat{r}_{k}(a,\boldsymbol{\psi})+\operatorname{conf}_{1,k}(a,\boldsymbol{\psi})$
and $\widehat{\mathcal{I}}_{k}=\operatorname{\arg\max}_{\mathcal{I}\in\mathcal{P}_{\leq m}(\mathcal{D})}\;\widehat{V}_{k}(\mathcal{I})$.
The pseudocode for the Sim-OOS is given in Algorithm~\ref{alg:PSim-OOS}. It can be easily shown that the computational complexity of the Sim-OOS
algorithm for $T$ instances is $\mathcal{O}\left(A\operatorname{poly}(\Psi_{tot})\log T\right)$.
\vspace{-0.1in}
\subsection{Regret Bounds for the Sim-OOS algorithm}
\vspace{-0.1in}
In this subsection, we provide distribution-independent regret bounds
for the Sim-OOS algorithm. Let $\boldsymbol{\psi}_{\text{tot}}=\sum_{\mathcal{I}\in\mathcal{P}_{\leq m}(\mathcal{D})}|\Psi^{+}(\mathcal{I})|$
denote the number of all possible states (all possible results
from at most $m$ distinct medical tests).

\begin{theorem} \label{thrm:Sim-OOSregret2} Suppose $\beta =1$. For any $0<\delta<1$,
set 
$$\operatorname{conf}_{1}(n,t)=\min\left(1,\sqrt{\frac{\log\left(20\Psi_{\text{tot}}At^{5}/\delta\right)}{2\max\left(1,n\right)}}\right)$$ and
$$\operatorname{conf}_{2}(n,t)=\min\left(1,\sqrt{\frac{10\Psi_{\text{tot}}\log\left(4t/\delta\right)}{\max\left(1,n\right)}}\right).$$ Then, with probability at least $1-\delta$, the regret of the Sim-OOS satisfies 
\begin{equation}
\operatorname{Reg}_{T}^{\text{Sim-OOS}}=\mathcal{O}\left(\left(\sqrt{A}+\sqrt{|\mathcal{P}_{\leq m}(\mathcal{D})|}\right)\sqrt{\Psi_{\text{tot}}T\log\left(T/\delta\right)}\right).\notag
\end{equation}
\end{theorem}

The proof of Theorem \ref{thrm:Sim-OOSregret2} and all the other results
can be found in the supplementary material. The UCRL2 (\citet{jaksch2010near})
is designed for general MDP problems and achieves a regret of $\tilde{O}\left(\sqrt{\Psi_{\text{tot}}^{2}AT}\right)$.
Hence, these regret results are better than those obtained by UCRL2.
This is an important result since it demonstrates that the Sim-OOS can
effectively exploit the structure of our CMAB-CO problem to achieve
efficient regret bounds which scale better than these that can be
obtained for general MDP problems.

We illustrate this bound using the same example above. Suppose $|\mathcal{X}_{i}|=X$
for all $i\in\mathcal{D}$ and $m=D$. The upper bound given in Theorem
\ref{thrm:Sim-OOSregret2} is in the order of $\tilde{\mathcal{O}}\left(\sqrt{\sum_{m=1}^{D}X^{m}2^{D}T}+\sqrt{\sum_{m=1}^{D}X^{m}AT}\right)$.

The Sim-OOS algorithm performs well for smaller values of which is the
case in the medical setting, as it is for instance the case in breast
cancer screening, in which imaging tests are limited to a small set:
mammogram, MRI and ultrasound~(\citet{saslow2007american}). In this context, the observations are usually selected sequentially.  To address such settings, we next propose the Seq-OOS algorithm that
selects observations sequentially.
\section{Multi-armed Bandits with Sequential Costly Observations} \label{sec:sequential}

\subsection{Problem Formalism}
Our current setting assumes that decision-maker makes all the observations simultaneously. If the decision-maker is allowed to make observations sequentially, she can use the partial state from already selected observations to inform the selection of future observations. For example, in the medical settings, although a positive result in a medical test is usually followed by additional medical test for validity, a negative result in a medical test is not usually followed by additional medical tests. Since any resulting simultaneous observation policy can be achieved by a sequential observation policy, the oracle defined with sequential observations achieves higher expected reward than that with simultaneous observations. At each time $t$, the following sequence of events is taking place: 
\begin{enumerate}[i]
\item The decision-maker has initially no observations. In phase $0$, we denote the empty partial state as $\psi_{0,t} = \psi_0$ where $\operatorname{dom}(\psi_0) = \emptyset$.
\item At each phase $l \in \mathcal{L} = \{1,\ldots,m \}$, if the partial state is $\psi_{l,t}$ and observation $i_{l,t} \in \left(\mathcal{D} \setminus \operatorname{dom}(\psi_{l,t})\right) \cup \emptyset$ is made, the resulting partial state is $\psi_{l+1,t}$ where $\psi_{l+1, t} =  \psi_{l,t} \cup \left( i_{l,t}, \phi_t(i_{l,t})   \right)$ if $i_{l,t} \neq \emptyset$ and $\psi_{l+1, t} =  \psi_{l,t}$ otherwise.
\item The decision-maker takes an action $a_t$ when either observation $i_{l,t} = \emptyset$ is made or the final phase $m$ is reached and observes a random reward $r_t$. 
\end{enumerate}

Let $\Psi^{+}(\psi,i)$ be the set of resulting partial state when observation $i$ is made at previous partial state of $\psi$, i.e., $\Psi^{+}(\psi,i) =\{\psi' : \exists x, \psi' = \psi \cup (i, x)  \}$. In this section, we define $p(\psi' | \psi, i)$ as the probability of resulting partial state $\psi'$ when the observation $i$ is made at previous partial state of $\psi$, which is referred to as \textit{partial state transition probability}. For all $\psi' \in \Psi^{+}(\psi, i)$, the partial state transition probability is defined as $p(\psi' | \psi, i) = \Pr(\Phi(i) = \psi'(i) | \Phi \sim \psi)$   if $i \in \mathcal{D} \setminus \operatorname{dom}(\psi)$ and $p(\psi' | \psi, i) = 0$ otherwise. In the medical example, this is the probability of observing test $i$'s result as $\psi'(i)$ given the previous test results (records) $\psi$. We define $p(\psi | \psi, \emptyset) = 1$ and $p(\psi' | \psi, \emptyset) = 0$ for all $\psi' \neq \psi$. Let $\boldsymbol{P} = \left[ p(\psi' | \psi, i) \right]$ denote partial state transition probability matrix.

A sequential policy $\pi = \{g, h \}$ consists of observation function $g$ and action function $h$ where $g : \Psi \rightarrow \mathcal{D} \cup \emptyset$ and $h : \Psi \rightarrow \mathcal{A}$ (e.g., $g(\psi)$ refers to the next medical test applied on a patient with previous records (test results) $\psi$ and $h(\psi)$ refers to treatment recommendation for a patient with previous records(test results) $\psi$). A sequential policy $\pi = \{g, h \}$ works as follows. Decision-maker keeps making observations $g(\psi)$ until either $m$ observations are made or an empty observation $g(\psi) = \emptyset$ is picked and takes an action $h(\psi)$ in a terminal state $\psi$ where terminal partial states of policy $\pi$ is the state with either cardinality $m$ or with $g(\psi) = \emptyset$. 

We illustrate these definitions in a medical example. Assume that there are $2$ different tests with possible outcomes of positive $(+)$ and negative $(-)$ result and $3$ different possible treatments. Suppose that a sequential policy $\pi = (g,h)$ with $g(\emptyset) = \{ 1\}, g(\{ (1, +)\}) = \{2 \}, g(\{ (1, -)\}) = \emptyset$, $h(\{ (1,+), (2,+) \}) = a_1$, $h(\{ (1,+), (2,-) \}) = a_2, h(\{ (1,-) \}) = a_3$. Basically, this policy initially picks the medical test $1$ for all patients ($g(\emptyset) = \{ 1\}$). If the result of the medical test $1$ is positive $(+)$, the policy picks medical test $2$  ($g(\{ (1, +)\}) = \{2 \}$). On the other hand, if the result of medical test $1$ is negative $(-)$, the policy does not make any additional test. In this example, terminal partial states of policy $\pi$ are $\psi_3, \psi_4, \psi_5$. 

Given a sequential policy $\pi$, let $\psi_l$ denote the random partial state in phase $l$ and $c_l = c_{g(\psi_l)}$ denote the random cost in phase $l$ by making observation $g(\psi_l)$. Note that $c_l$ is random since partial state in phase $l$ is random. Similarly, let $r_m$ denote random reward revealed by taking action $a_m = h(\psi_m)$ in terminal partial state. Then, for each sequential policy $\pi = (g,h)$, we define a value function for $l=0,\ldots, m$:
\begin{eqnarray}
F_l^{\pi}(\psi) = \mathbb{E}\Big[ \beta r_m - \sum_{\tau = l}^{m-1} c_{\tau}  \bigg| \psi_l = \psi , \pi \Big], 
\end{eqnarray}
where expectation is taken with respect to randomness of the states and rewards. In the terminal phase, we define value function as $F_m^{\pi}(\psi) = \bar{r}(h(\psi), \psi)$. The optimal value function is defined by $F^{*}_l(\psi) = \sup_{\pi} F_l^{\pi}(\psi)$. A policy $\pi^{*}$ is said to be optimal if $F_0^{\pi^{*}}(\psi) = F^{*}_0(\psi)$. It is also useful to define partial state-observation optimal value function for $l = 0, \ldots, m-1$ : 
\begin{eqnarray*}
 Q^{*}_l(\psi, i) &=& \mathbb{E}\left[ -c_i + F^{*}_{l+1}(\psi_{l+1}) | \psi_l = \psi, i_l = i \right]  \\
&=& -c_i + \sum_{\psi' \in \Psi^{+}(\psi, i)} p(\psi' | \psi, i) F^{*}_{l+1}(\psi').  
\end{eqnarray*}
A sequential policy $\pi^{*} = (g^{*},h^{*})$ is optimal if and only if  $g^{*}(\psi) = \operatorname{\arg\max}_{i \in \left(\mathcal{D} \cup \emptyset \right) } Q^{*}_{|\operatorname{dom}(\psi)|}(\psi, i)$, $h^{*}(\psi) = \operatorname{\arg\max}_{a \in \mathcal{A}} \bar{r}(a, \psi)$. 

Consider a sequential learning algorithm $\pi_{1:T} = (g_t, h_t)_{t = 1}^{T}$. The algorithm makes observation $i_{l,t} = g(\psi_{l,t})$ and realizes a cost $c_{l,t}$ in phase $l$ of time $t$ and then selects action $a_t = h_t(\psi_{m,t})$ and realizes a random reward $r_t$, which realizes a reward of $r_t - \sum_{l=0}^{m-1} c_{l,t}$. To quantify the performance of sequential learning algorithm, we define cumulative  regret of sequential learning algorithm $\pi_{1:T}$ up to time $T$ as
\begin{align}
\operatorname{Reg}_{T}^{\pi_{1:T}} = T F^{*}_0(\psi_0) - \sum_{t=1}^T \left( r_t - \sum_{l=0}^{m-1} c_{l,t} \right) \notag 
\end{align}
where $\psi_0 = \emptyset$ denotes empty state. In the next subsection, we propose a sequential learning algorithm, which aims to minimize regret. 

\subsection{Sequential Optimistic Observation Selection (Seq-OOS)}
In addition to observation sets that are tracked by  Sim-OOS, Seq-OOS keeps track of the following sets at each round $k$ : $\mathcal{E}_k(\psi, i) = \{\tau < t_k : \exists l \in \mathcal{L}, \; \psi_{l,\tau} = \psi, i_{l,t} =i \}$,  $\mathcal{E}_k(\psi, i, \psi') = \{\tau < t_k : \exists l \in \mathcal{L}, \; \psi_{l,\tau} = \psi, i_{l,\tau} =i, \psi_{l +1,\tau}  = \psi'\}$.
Let $N_k(\psi, i) = |\mathcal{E}_k(\psi, i)|$ and $N_k(\psi, i, \psi') = |\mathcal{E}_k(\psi, i, \psi')|$. In addition to these counters, we also keep counters of visits in partial state-action pairs and state-observation pairs in a particular round $k$. Let $\nu_k(\psi, i)$ denote the number of times observation $i$ is made when partial state $\psi$ is realized in round $k$. We can express the estimated transition probabilities as $\hat{p}_k(\psi'|\psi, i)= \frac{N_k(\psi, i, \psi')}{N_k(\psi, i)}$, provided that $N_k(\psi, i) >0$. 

The Seq-OOS works in rounds $k = 1,\ldots$. In the beginning of round $k$ ($t_k$ denotes time of beginning of round $k$), the Seq-OOS solves Optimistic Dynamic Programming (ODP), which takes the estimates $\hat{\boldsymbol{P}}_k =\left[\hat{p}_k( \psi'| \psi, i)\right]$ and $\hat{\boldsymbol{R}}_k = \left[\hat{r}_k(a, \psi)\right]$ as an input and outputs a policy $\pi_k$. The ODP first orders the partial states with respect to size of their domains. Let $\Psi_{l}$ denote partial states with $l$ observations, which is defined by $\Psi_l = \{\psi : |\operatorname{dom}(\psi)| = l\}$ (e.g., all possible results from $l$ distinct medical tests). Since the decision-maker is not allowed to make any more observations for any state $\psi \in \Psi_m$, estimated value of state $\psi$ is computed by $\hat{F}_{m,k}(\psi) = \max_{a \in \mathcal{A}} \hat{r}_k(a, \psi) + \operatorname{conf}_{1,k}(a, \psi)$ where $\operatorname{conf}_{1,k}(a, \psi)$ is the confidence interval for partial state-action pair in round $k$. The action and observation functions on partial state $\psi \in \Psi_m$ computed by ODP is given by $\hat{g}_k(\psi) = \emptyset$ and $\hat{h}_k(\psi) = \operatorname{\arg\max}_{a \in \mathcal{A}} \hat{r}_k(a, \psi) + \operatorname{conf}_{1,k}(a, \psi)$. After computing value and policy in partial states $\psi \in \Psi_m$, the ODP solves convex optimization problem to compute optimistic value function for each partial state-observation pair  $\psi \in \Psi_{m-1}$ and $i \in \mathcal{D} \setminus \operatorname{dom}(\psi)$. Let $\hat{Q}_{m-1,k}(\psi, i)$ denote optimistic value function for making observation $i$ in partial state $\psi$ in round $k$ of phase $m-1$, which is the solution of the following convex optimization problem :  
\begin{align}
& \underset{\left[\tilde{p}(\cdot | \psi, i)\right]}{\operatorname{maximize}} -c_i + \sum_{\psi' \in \Psi^{+}(\psi, i)} \tilde{p}(\psi' | \psi, i)\hat{F}_{m,k}(\psi') \notag \\
& \text{subject to} \;\; \sum_{\psi' \in \Psi^{+}(\psi, i)}| \tilde{p}(\psi' | \psi,i) - \hat{p}_k(\psi' | \psi,i) | \leq \operatorname{conf}_{2,k}(\psi, i), \notag \\ 
&\;\;\;\;\;\;\;\;\;\;\;\;\;\;\;\; \sum_{\psi' \in \Psi^{+}(\psi,i)} \tilde{p}( \psi' | \psi,i) =1. \label{eqn:optimization_2_1} 
\end{align}

Note that the variables ($\hat{F}_{m,k}(\psi')$) used in the convex optimization problem given in (\ref{eqn:optimization_2_1}) is computed in the previous step by the ODP. The optimistic value of the empty observation $\emptyset$ in partial state $\psi$ in round $k$ is computed by $\hat{Q}_{m-1, k}(\psi, \emptyset) = \max_{a \in \mathcal{A}} \beta \hat{r}_k(a, \psi) + \operatorname{conf}_{1, k}(a, \psi)$. Based on the optimistic value of partial state-observation pairs $\left[ \hat{Q}_{m,k}(\psi, i) \right]$, the ODP computes the optimistic value of partial state $\psi$ and action and observation function of partial state $\psi \in \Psi_{m-1}$ as $\hat{F}_{m-1, k}(\psi) = \max_{i \in \left(\mathcal{D} \setminus \operatorname{dom}(\psi) \right) \cup \emptyset } \hat{Q}_{m-1, k}(\psi, i)$, $\hat{h}_k(\psi) = \operatorname{\arg\max}_{a \in \mathcal{A}} \beta \hat{r}_k(a, \psi) + \operatorname{conf}_{1, k}(a, \psi)$, $\hat{g}_k(\psi) = \operatorname{\arg\max}_{i \in \left( \mathcal{D} \setminus \operatorname{dom}(\psi)\right) \cup \emptyset} \hat{Q}_{m-1,k}(\psi, i)$. These computations are repeated for $l = m-2, \ldots, 0$ to find the complete policy $\hat{\pi}_k$.

Given $\hat{\pi}_k = (\hat{g}_k ,\hat{h}_k)$, at each time $t$ of round $k$ ($t_{k-1} \leq t \leq t_k$), the Seq-OOS follows the policy $\hat{\pi}_k$. Basically, if the state at phase $l$ is $\psi_{l,t}$, the Seq-OOS decides to make the observation $i_{l,t} = \hat{g}_k(\psi_{l,t})$ and observes the state $\psi_{l+1,t}$. If the state is $\psi_{l,t}$ at phase $l < m$ and observation $i_{l,t} = \hat{g}_k(\psi_{l,t})$ computed by the ODP is empty set, i.e., $\hat{g}_k(\psi_{l,t}) = \emptyset$, then Seq-OOS takes action $\hat{h}_k(\psi_{l,t})$. If it is a terminal phase, i.e., $l=m$, Seq-OOS takes an action $\hat{h}_k(\psi_{m,t})$. 

\subsection{Regret Bounds of the Seq-OOS}
The analysis of the regret of the Seq-OOS  exhibits similarities to the analysis of the regret of the Sim-OOS. The Seq-OOS has at most $m+1$ phases in which it makes observations sequentially followed by an action while Sim-OOS has $2$ phases in which it makes simultaneous observations at once followed by an action. The difference is that we need to decompose the regret of the Seq-OOS into regret due to phases with suboptimal observations and regret due to suboptimal actions. Let $\Psi_{\max} = \max_{\psi} \max_{i \in \mathcal{D}} |\Psi^{+}(\psi, i)|$. The next theorem bounds the distribution-independent regret. 

\begin{theorem} \label{thrm:SSim-OOSregret2} Suppose $\beta = 1$. For $0< \delta < 1$, set $$\operatorname{conf}_1(n,t) = \min\left(1, \sqrt{\frac{\log\left( 20 \Psi_{\text{tot}} A t^5/\delta \right)}{2 \max\left(1,n\right)}}\right)$$ and $$\operatorname{conf}_2(n,t) = \min\left(1, \sqrt{\frac{10 \Psi_{\max} \log\left( 4 D \Psi_{\text{tot}} t/ \delta \right)}{ \max\left(1,n\right)}}\right).$$Then, with probability at least $1 - \delta$, regret of the Seq-OOS satisfies $$\operatorname{Reg}_T^{\text{Seq-OOS}} = \mathcal{O}\left(  \left(m \sqrt{\Psi_{\text{max}} D} + \sqrt{A}\right) \sqrt{\Psi_{\text{tot}}  T \log \left(T/ \delta \right)} \right)$$. 
\end{theorem}

The difference in the regret bounds of Sim-OOS and Seq-OOS is because Sim-OOS estimates the observation probabilities $p(\psi)$ for each $\psi \in \Psi$ whereas Seq-OOS estimates observation transition probabilities $p(\cdot | \psi, i)$ for each $\psi \in \Psi$ and $i \in \mathcal{D}$. 

Now, we illustrate and compare the regret bounds on our algorithms. Suppose that $|\mathcal{X}_i| = X$ for all $i \in \mathcal{D}$ and $m = D$. In this case, we have the distribution independent regret of $O\left( 2^D \sqrt{ A X^D \log T / \delta }\right)$ for Sim-OOS and $\left( D \sqrt{D 2^D X^{D+1} A T \log T/ \delta}\right)$ for Seq-OOS with probability at least $1 - \delta$. Our algorithms become computationally feasible when $X^D$ is small.

\begin{figure}[h]
    \centering
        \includegraphics[width=0.48\textwidth]{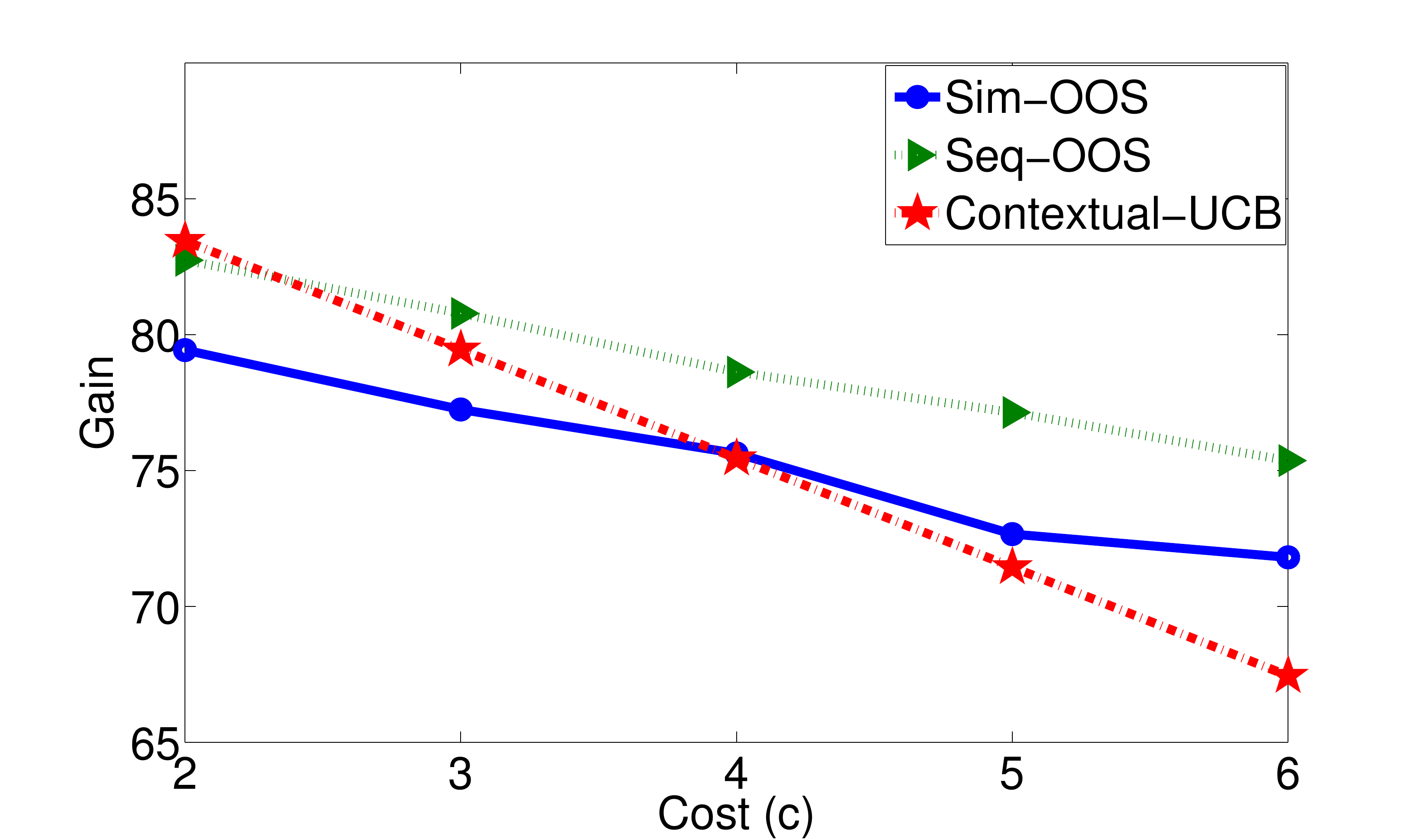}      
    ~ 
      \label{fig:m_3}
    \caption{Comparison of Sim-OOS, Seq-OOS and All-Context UCB}
\end{figure}
\section{Illustrative Results} \label{sec:illustrative}
We evaluate the Sim-OOS and Seq-OOS on a dataset of 10,000 records of breast cancer patients participating in the National Surgical Adjuvant Breast and Bowel Project (NSABP) by~["removed for anonymous submission"]. Each instance consists of the following information about the patient: age, estrogen receptor, tumor stage, WHO score. The treatment is a choice among four chemotherapy regimes AC, ACT, CAF, CEF. The outcomes for these regimens were derived based on 32 references from PubMed Clinical Queries; this is a medically accepted procedure. Hence, the data contains the feature vector and all derived outcomes for each treatment. The details are given in~["removed for anonymous submission"]. We generate $200000$ instances by randomly selecting a sample from the breast cancer dataset. In each instance, we set the observations as $\mathcal{D} =\{ \text{age}, \text{estrogen receptor}, \text{tumor stage}, {WHO Score}\}$, and the rewards as $1$ if the treatment with the highest outcome is given to the patient and $0$ otherwise. For the experimental results, we set $\beta = 100$ and $m=3$.

We compare Sim-OOS and Seq-OOS algorithms with a contextual bandit algorithm that observes realization of all observation states $\phi$ by paying cost of $\sum_{i=1}^D c_i$, referred to as Contextual-UCB. We define the following metric of \textit{Gain} of our algorithms ,which make observations $\mathcal{I}_t$ and receives reward of $r_t$ by taking action $a_t$ at each time $t$, over $T$ time steps by $\operatorname{Gain} = \frac{1}{T} \sum_{t=1}^T \left[ \beta r_t - \sum_{i \in \mathcal{I}_t} c_i \right]$. 

\textbf{Performance of the Sim-OOS and Seq-OOS with Different Costs:} We consider that the cost of each observation $c_i=c$. We illustrate gain of Sim-OOS, Seq-OOS and Contextual-UCB algorithms for increasing values of cost $c$. As Figure 1 illustrate, the gain of the Sim-OOS and Seq-OOS algorithm decreases as the observation cost increases. However, it should be noted that these algorithms learn the best simultaneous and sequential policies while simultaneously taking actions irrespective of the costs of observation. These figures show that when the observation cost is increasing, the Sim-OOS and Seq-OOS achieves better gains than Contextual-UCB by observing less information, hence paying less cost. Therefore, the slope of the gain-cost curve of the Sim-OOS and Seq-OOS illustrated in Figure 1 decreases as the observation cost increases. 
\section{Conclusions}\label{sec:conclusion}
\vspace{-0.1in}

In this paper, we introduced the novel, yet ubiquitous problem of
contextual MAB with costly observations: selecting what information
(contexts) to observe to inform the decision making process. To address
this problem, we developed two different algorithms: Sim-OOS and Seq-OOS, 
and prove that these algorithms achieve distribution-independent regret bounds that are sublinear
in time. Future work will be dedicated to exploring algorithms with regret bounds that are polynomial on the number of observations. 

\bibliographystyle{abbrvnat}
\bibliography{nips_1}

\begin{thebibliography}{18}
\providecommand{\natexlab}[1]{#1}
\providecommand{\url}[1]{\texttt{#1}}
\expandafter\ifx\csname urlstyle\endcsname\relax
  \providecommand{\doi}[1]{doi: #1}\else
  \providecommand{\doi}{doi: \begingroup \urlstyle{rm}\Url}\fi

\bibitem[Auer et~al.(2002)Auer, Cesa-Bianchi, and Fischer]{auer2002a}
P.~Auer, N.~Cesa-Bianchi, and P.~Fischer.
\newblock Finite-time analysis of the multi-armed bandit problem.
\newblock \emph{Machine Learning}, 47:\penalty0 235--256, 2002.

\bibitem[Boyd and Vandenberghe(2004)]{boyd2004convex}
S.~Boyd and L.~Vandenberghe.
\newblock \emph{Convex optimization}.
\newblock Cambridge university press, 2004.

\bibitem[Campbel and Kracaw(1980)]{campbel1980information}
T.~S. Campbel and W.~A. Kracaw.
\newblock Information production, market signalling, and the theory of
  financial intermediation.
\newblock \emph{The Journal of Finance}, 35\penalty0 (4):\penalty0 863--882,
  1980.

\bibitem[Cesa~Bianchi et~al.(2011)Cesa~Bianchi, Shalev~Shwartz, and
  Shamir]{cesa2011efficient}
N.~Cesa~Bianchi, S.~Shalev~Shwartz, and O.~Shamir.
\newblock Efficient learning with partially observed attributes.
\newblock \emph{The Journal of Machine Learning Research}, 12:\penalty0
  2857--2878, 2011.

\bibitem[Chemmanur(1993)]{chemmanur1993pricing}
T.~J. Chemmanur.
\newblock The pricing of initial public offerings: A dynamic model with
  information production.
\newblock \emph{The Journal of Finance}, 48\penalty0 (1):\penalty0 285--304,
  1993.

\bibitem[Chu et~al.(2011)Chu, Li, Reyzin, and Schapire]{chu2011contextual}
W.~Chu, L.~Li, L.~Reyzin, and R.~E. Schapire.
\newblock Contextual bandits with linear payoff functions.
\newblock In \emph{International Conference on Artificial Intelligence and
  Statistics}, pages 208--214, 2011.

\bibitem[Dudik et~al.(2011)Dudik, Hsu, Kale, Karampatziakis, Langford, Reyzin,
  and Zhang]{dudik2011efficient}
M.~Dudik, D.~Hsu, S.~Kale, N.~Karampatziakis, J.~Langford, L.~Reyzin, and
  T.~Zhang.
\newblock Efficient optimal learning for contextual bandits.
\newblock \emph{arXiv preprint arXiv:1106.2369}, 2011.

\bibitem[Golovin and Krause(2010)]{golovin2010adaptive}
D.~Golovin and A.~Krause.
\newblock Adaptive submodularity: A new approach to active learning and
  stochastic optimization.
\newblock In \emph{COLT}, pages 333--345, 2010.

\bibitem[Hazan and Koren(2012)]{hazan2012linear}
E.~Hazan and T.~Koren.
\newblock Linear regression with limited observation.
\newblock In \emph{Proc. 29th Int. Conf. on Machine Learning}, pages 807--814,
  2012.

\bibitem[Jaksch et~al.(2010)Jaksch, Ortner, and Auer]{jaksch2010near}
T.~Jaksch, R.~Ortner, and P.~Auer.
\newblock Near-optimal regret bounds for reinforcement learning.
\newblock \emph{Journal of Machine Learning Research}, 11:\penalty0 1563--1600,
  2010.

\bibitem[Langford and Zhang(2007)]{langford2007epoch}
J.~Langford and T.~Zhang.
\newblock The epoch-greedy algorithm for contextual multi-armed bandits.
\newblock \emph{Advances in Neural Information Processing Systems (NIPS)},
  20:\penalty0 1096--1103, 2007.

\bibitem[Lu et~al.(2010)Lu, P{\'a}l, and P{\'a}l]{lu2010contextual}
T.~Lu, D.~P{\'a}l, and M.~P{\'a}l.
\newblock Contextual multi-armed bandits.
\newblock In \emph{International Conference on Artificial Intelligence and
  Statistics}, pages 485--492, 2010.

\bibitem[Ortner and Auer(2007)]{ortner2007logarithmic}
P.~Ortner and R.~Auer.
\newblock Logarithmic online regret bounds for undiscounted reinforcement
  learning.
\newblock In \emph{Advances in Neural Information Processing Systems}, 2007.

\bibitem[Osband et~al.(2016)Osband, Van~Roy, and Wen]{osband2014generalization}
I.~Osband, B.~Van~Roy, and Z.~Wen.
\newblock Generalization and exploration via randomized value functions.
\newblock In \emph{International Conference on Machine Learning}, 2016.

\bibitem[Saslow et~al.(2007)Saslow, Boetes, Burke, Harms, Leach, Lehman,
  Morris, Pisano, Schnall, Sener, et~al.]{saslow2007american}
D.~Saslow, C.~Boetes, W.~Burke, S.~Harms, M.~O. Leach, C.~D. Lehman, E.~Morris,
  E.~Pisano, M.~Schnall, S.~Sener, et~al.
\newblock American cancer society guidelines for breast screening with mri as
  an adjunct to mammography.
\newblock \emph{CA: a cancer journal for clinicians}, 57\penalty0 (2):\penalty0
  75--89, 2007.

\bibitem[Slivkins(2011)]{slivkins2011contextual}
A.~Slivkins.
\newblock Contextual bandits with similarity information.
\newblock In \emph{24th Annual Conference On Learning Theory}, 2011.

\bibitem[Tekin and Van Der~Schaar(2014)]{tekin2014discovering}
C.~Tekin and M.~Van Der~Schaar.
\newblock Discovering, learning and exploiting relevance.
\newblock In \emph{Advances in Neural Information Processing Systems}, pages
  1233--1241, 2014.

\bibitem[Zolghadr et~al.(2013)Zolghadr, Bart{\'o}k, Greiner, Gy{\"o}rgy, and
  Szepesv{\'a}ri]{zolghadr2013online}
N.~Zolghadr, G.~Bart{\'o}k, R.~Greiner, A.~Gy{\"o}rgy, and C.~Szepesv{\'a}ri.
\newblock Online learning with costly features and labels.
\newblock In \emph{Advances in Neural Information Processing Systems (NIPS)},
  pages 1241--1249, 2013.

\end{thebibliography}
 
\end{document}